# Automating rule generation for grammar checkers

Marcin Miłkowski

Polish Academy of Sciences

## 1. Introduction

In this paper, I describe several approaches to automatic or semi-automatic development of symbolic rules for grammar checkers from the information contained in corpora. The rules obtained this way are an important addition to manually-created rules that seem to dominate in rule-based checkers. However, the manual process of creation of rules is costly, time-consuming and error-prone. It seems therefore advisable to use machine-learning algorithms to create the rules automatically or semi-automatically. The results obtained seem to corroborate our initial hypothesis that symbolic machine learning algorithms can be useful for acquiring new rules for grammar checking. It turns out, however, that for practical uses, error corpora cannot be the sole source of information used in grammar checking. We suggest therefore that only by using different approaches, grammar-checkers, or more generally, computer-aided proofreading tools, will be able to cover most frequent and severe mistakes and avoid false alarms that seem to distract users.

In what follows, I will show how Transformation-Based Learning (TBL) algorithms may be used to acquire rules. Before doing that, I will discuss the pros and cons of three approaches to creating rules and show the need to make use of them all in a successful grammar-checking tool. The results obtained seem to suggest that the machine-learning approach is actually fruitful, and I will point to some future work related to the reported research.

## 2. Three approaches to rule-based grammar checking

It could seem that a *pure empirical approach* is the most advisable way to create rules for grammar checkers. However, it must rely on error corpora, and these are hard to find and costly. Although it is possible to automatically bootstrap the development of such corpora (see Miłkowski 2008), they are still not sufficient to cover all variety of important errors in the language on the proper level of generality. Direct use of smaller error corpora does not seem to lead to useful results (see below), and they must be accompanied with a "clean" corpus that contains no or only minor errors. The rules acquired this way, however, can be a useful addition to the existing base of rules.

*Pure a priori approach* as based on existing normative prescriptions, such as usage dictionaries, style guides, grammar textbooks etc. offers two significant advantages over the pure empirical approach. First, the justification of the rules seems to be more reliable. It is the job of lexicographers to check if the advice offered is accurate and authoritative, and the authors of grammar-checkers can simply rely on it. Moreover, if the grammar checker is to impose style guidelines, then it is doubtful that any empirical corpus would embody them beforehand. For example, if the style guide requires that dates be written only in a recommended format, it is easier to define the standard format and use the corpus only to find out what deviations seems to occur most frequently. This is why for some uses, the a priori approach cannot be replaced with pure empirical one.

For larger-scale projects, however, pure a priori approach has some disadvantages: it lags behind linguistic development, as dictionaries tend to be outdated (especially the printed ones). In other words, some errors listed in the dictionaries may no longer exist in the language, and some existing errors may be missing from the dictionary. Both factors can negatively affect the recall of grammar-checking rules. For example, Polish dictionaries of usage usually advise against using cliches of Soviet propaganda, including especially bureaucratic borrowings from Russian. It turns out that those cliches are almost extinct nowadays; it makes therefore little sense to develop rules detecting them as contemporary speakers are not likely to use the cliches anyway. At the same time, with the wide adoption of computers for word processing, spell-checkers seem to introduce specific kinds of errors into the text. One of the basic functions in spell-checkers is detecting run-on words but when implemented carelessly, it seems to introduce errors in Polish neologisms (for example, the word "technohumanizm", which is a correct productive neologism, missing in spelling dictionaries, could be automatically corrected to "techno humanizm", which is incorrect but composed of two correct dictionary words). This kind of errors is however completely missing from usage dictionaries and similar resources, so by relying completely on them, we would fail to detect it.

There is also another factor that makes a priori approach hard in practical applications. Culturally-transmitted normative prescriptions for linguistic behavior are vague and tend to lack contextual information needed to disambiguate between correct cases and incorrect ones, even if some examples are given. This is true of usage dictionaries, spelling dictionaries, and grammar textbooks. Moreover, as some prescriptions are on the semantic level, which is not fully reducible to the surface level on which most grammar checkers operate, they need to be (sometimes painfully) adjusted to the Procrustean bed of what is expressible on the surface level. As a result, even authors with a background in linguistics need process a corpus with the newly created rules to make sure that they are not resulting in a very high number of false alarms. Yet, as our research shows, the false alarms can also result from automatic acquisition of rules (which is shown by precision levels of 30%), so the step of processing corpora with the rules and refining them seems unavoidable.[1]

Lastly, minority languages can lack authoritative sources altogether, so in some cases such resources cannot be reused at all. As one of the goals of our research project was to alleviate the process of adding new rules to an open-source proofreading tool, LanguageTool (for a general description, see Naber 2003), and we wanted to include minority languages among the ones supported, using reliable dictionaries could not be a requirement.

---

1 However, some automatically-acquired rules can be discarded automatically if they seem to result in a very high number of matches in a substantial corpus that is considered to be relatively free of errors.

It is possible to use the information from both worlds: machine learning algorithms with some existing language resources, such as authoritative usage guides. Although the quality of some automatically-developed rules may be lower in terms of their precision, we may restrict ourselves to the best ones and refine others manually. *The mixed approach* uses therefore both corpora to create rules using real-world data and some prescriptions that lead the process of acquisition of rules.

## 3. Machine learning of rules

One of the machine-learning techniques that was used successfully in the mixed approach is, as we already mentioned, Transformation-Based Learning (see Mangu & Brill 1997) but it can also be used in the pure empirical approach, as shown in the section 5. TBL is of course just one of the available machine-learning algorithms that may acquire rules. It is also quite straightforward to use statistical machine translation (SMT), which is already applied in automatic post-editing (Simard et al. 2007) and could also be quite easily adopted to add missing words in the text (like determiners or missing parts of common phrases). It might be used for ESL grammar checking (Gamon et al. 2008). Here, I will focus on TBL, as it creates rules that can be easily translated into LanguageTool symbolic notation (constructs admissible in TBL rules are a subset of what is expressible in LT notation). One of the interesting properties of TBL is that the training corpus can be relatively small to achieve good results, contrary to most SMT-based methods that are data-intensive.

TBL is most well-known for its application in the Brill tagger (Brill 1995), however, it may be used in a wide variety of NLP tasks that involve classification. The algorithm starts with an initial classification, and tries to find transformations from initial classes to the correct ones. In the case of part-of-speech (POS) tagging, it transforms most frequent POS tags, which were applied first, into correct ones, which are taken from a manually-annotated corpus. To constrain the search space of possible transformations, the TBL algorithm uses a finite list of admissible transformation templates that usually include features (POS tags, surface tokens, lemmas) that precede the feature under consideration or follow it.

The application of TBL to the error-detection task is straightforward: we need to find the transformation from incorrect tokens into the correct ones. Brill used TBL to find commonly confused words in English, by simply substituting POS tags in the POS tagging task with confused words.[2] The confused words create (at least two-element) confusion sets such as {*than*, *then*}, {*its, it's*}, {*there, their*}, {*advice, advise*}. For any member of the confusion set, the original word was replaced with other(s) in a clean corpus, and the original word was moved into the place of the correct feature. In a familiar column format of Brill tagger, this could look like this (assuming that POS tags were used in the clean corpus):[3]

```
NN      advice      advise
```

Next, the standard learning procedure is run. It results in a ranked list of transformation rules. The original paper of Mangu and Brill used a dictionary-based word list, but TBL can be used with confusion sets created in a different way, using additional corpora or without them.

For example, one could use artificially created errors in a clean corpus (Sjöberg & Knuttson 2005). Such approach has been taken in transformation-based learning for punctuation correction in Danish (Hardt 2001). Note that artificial errors should not be just randomly created; otherwise, the rules created will have no recall on normal text, as nobody will actually make mistakes detected by those rules (as Miłkowski 2008 observes, this was the case of automatically created autocorrection rules for early versions of Polish OpenOffice.org). So, the search space of possible errors should correspond to the space of frequent errors typical of the data-entry method used (OCR, standard keyboard, mobile phone on-screen keyboard...). For example, if the error corpus is to contain frequent typos and the grammar checker will operate in a word processor, then the errors should be selected based on distances between letters in the most popular keyboard layout (*c* replaced with *v* more frequently than with *p* in the case of QWERTY keyboards). If the proofreading tool is to be used in a digitalization project on OCRed text, the typos should be selected based on graphical character similarity in typical typefaces (*l* replaced with *1* more frequently than with *k*). Yet, in all those cases the simulated errors are only mechanical rather than cognitive mistakes that result from insufficient linguistic competence. It is much harder to artificially simulate errors involving confused nouns, non-standard versions of idioms, or false-friends in a translated text without any additional language resources.

---

[2] The list of words was simply excerpted from a dictionary (Flexner 1983).

[3] Note that the POS tag used is the one that corresponds to the confused word.

An obvious source of confusion set is an error corpus. Approaches based on reusing error corpora were often discarded as non-realistic, as creating such corpora is costly. Yet, there are ways to automate building such corpora by observing the frequency of user revisions to the text (Miłkowski 2008). Frequent minor edits tend to be related to linguistic mistakes and textual conventions. They have been used to learn transformation rules that automatically correct machine-translated text and improve its quality (Elming 2008).

It can be noted that for some purposes, automating rules for sticking to a linguistic convention may be desired. So, for example, if Wikipedia editors usually correct some non-adherence to the Wikipedia typographical standards, they could use a proofreading tool to scan the texts for non-compliance with those standards. However, in a general proofreading tool those conventional edits should be filtered out; either by averaging on many different revision histories from various sources (which is not yet realistic), or by manually filtering out such frequent revisions.

## 4. Approaches to learning from the error corpus

It might seem that one could simply use the corpus of errors (or revisions) directly. But rules created from a corpus tend to have high recall with low precision: there are simply not enough examples for members of the same confusion set to discriminate the cases during learning. Here are the highest-ranked rules created by fnTBL toolkit (Ngai & Florian 2001) on a Holbrook corpus used directly (Holbrook 1964):[4]

```
GOOD:21 BAD:0 SCORE:21 RULE: pos_0=Jame word_0=Jame =>
pos=James
GOOD:20 BAD:0 SCORE:20 RULE: pos_0=dont word_0=dont =>
pos=don't
GOOD:10 BAD:0 SCORE:10 RULE: pos_0=two pos_1=the
word_0=two word_1=the => pos=to
GOOD:10 BAD:0 SCORE:10 RULE: pos_0=yow word_0=yow =>
pos=you
GOOD:10 BAD:0 SCORE:10 RULE: pos_0=thats word_0=thats =>
pos=that's
GOOD:10 BAD:0 SCORE:10 RULE: pos_0=oclock word_0=oclock
=> pos=o'clock
```

As is easily seen, some of these rules are hardly useful in a normal context. All occurrences of "Jame" are unconditionally transformed into "James" (there are more examples like that, for example "cave" being replaced with "gave"). Moreover, it turns out that the third rule is based on a mistaken tagging[5] in the original corpus (the ERR *tag* contains the suggested correction as the contents of *targ* attribute):

> *15 NOVEMBER 1960 The Redex Jack O'Malley: Boss George Green 2 Head John Young 2 head went <ERR targ=***two***> to </ERR> the <ERR targ=pictures> pictyres </ERR> and hit the <ERR targ=manager> maneger </ERR> and he went down the <ERR targ=stairs> stars </ERR> and hit a <ERR targ=woman> women </ERR> and <ERR targ=she> he </ERR> fainted We home.*

---

4   I left the original names of features from the Brill tagger to stress the connection between both tasks.
5   During writing of this paper, the error has been fixed in the edition of Holbrook corpus maintained in computer-readable form by Roger Miton.

This shows that error corpora, which tend to be small, do not offer a sufficient number of examples of features to be learned. Moreover, a slight error in tagging can lead to surprisingly bad results. This is again a question of scale, as in a huge error corpus, a single tagging error could become irrelevant.

However, there is a way to use small corpora as seeds to create artificial errors in a large clean corpus. In other words, we need to extract only confusion sets ({*two*, *to*}, {*pictures*, *pictyres*}...) from the error corpus, possibly filter non-dictionary words as they are already flagged by standard spelling checkers, and apply confusion sets to the clean corpus. A bigger corpus will have more examples, and that results with higher precision. Let us look at the result of learning now:

```
GOOD:4845 BAD:0 SCORE:4845 RULE: pos_0=end word:[-3,-1]=, => pos=and
GOOD:1443 BAD:0 SCORE:1443 RULE: pos_0=end word:[1,3]=PRP => pos=and
GOOD:839 BAD:0 SCORE:839 RULE: pos_0=end pos:[1,3]=. => pos=and
GOOD:681 BAD:0 SCORE:681 RULE: pos_0=end word:[1,3]=DT => pos=and
GOOD:408 BAD:0 SCORE:408 RULE: pos_0=end word:[-3,-1]=DT => pos=and
GOOD:354 BAD:0 SCORE:354 RULE: pos_0=end word:[1,3]=, => pos=and
```

The rules concerning the confusion of "end" and "and" make now much more sense: "end" most of the time requires a determiner, so it is quite improbable that it will occur straight after a comma or before a determiner. The rule that suggests that "end" is incorrect before a dot (third one) might be however incorrect and result from a too small learning sample. The Table 1 contrasts results achieved using the (1) naïve learning method and (2) introducing extracted confusion sets to a clean corpus (*mixed learning* for short). The resulting rules were applied to the Brown corpus (Francis & Kucera 1964).

| Corpus type | Measure | Naïve learning | Mixed learning |
|---|---|---|---|
| Training corpus | Recall | 20.34% | 99.43% |
| | Precision | 95.11% | 99.97% |
| Brown corpus | Recall | 0.00% | 100.00% |
| | Precision | 0.00% | 38.00% |

The results obtained suggest that rules created directly from Holbrook corpus by naively using it for learning have no practical value for any data sets that were not included in the training. In other words, using the naïve approach, especially with a small error corpus, makes no sense; only the mixed approach offers some interesting results but precision is also relatively low. It means that only some rules could be usefully used for grammar checking.

Yet, Holbrook corpus seems to be too tiny for almost any machine learning task. This is why we evaluated the strategies on a much larger error corpus, i.e., the one we previously created from the whole Polish Wikipedia revision history (Miłkowski 2008). As a baseline, we used the Frequency Dictionary of Polish corpus, which is comparable to Brown (Kurcz et. al 1974-1977). The results are:

| Corpus type | Measure | Naïve learning | Mixed learning |
|---|---|---|---|
| Training corpus | Recall | 33.60% | 99.99% |
| | Precision | 80.14% | 100.00% |
| Frequency Dictionary Corpus | Recall | 100.00% | 100.00% |
| | Precision | 0.34% | 34.65% |

As can be seen, much bigger a training corpus does not seem to change the situation: the naïve approach has very low precision, and detects almost any occurrence of a confused word as an error (hence, 100% recall). As before,

the mixed approach enhances precision significantly but the rules still need human evaluation.[6]

One more thing deserves a mention. The original idea of applying TBL to detect errors focused only on confused words. It is also possible to create rules to detect missing words; the transformation words would simply substitute a missing word for the artificial token "NULL". We tested the idea by removing the word "się", which is required by some reflexive verbs in Polish; the resulting rules cite almost only reflexive verbs. In the case of words that were written separately but sometimes need to be spelled together ("no body" vs. "nobody" in English), the confusion set can simply contain an artificial word "no-body", so that mapping between subsequent features is retained. As long as the error form can be brought into a form of transforming a single feature into a target one (incorrect word into a correct word), TBL can be used. For more complex structural transformations, it could become less viable.

## 5. Adapting the rules to LanguageTool and future work

TBL rules, as noted, can be easily translated into the notation of declarative symbolic rules used in LanguageTool (the exact description of the notation is outside the scope of this paper). Rules are specified in XML files as patterns of tokens; tokens can be tested for multiple attributes (such as POS tags, lemmas and surface forms, and preceding white space) and relations (including conjunction, negation, exceptions, conditional skipping of tokens forward, as well as feature unification). The attributes can be specified as regular expressions. For that reason, the rules can be much more general than the ones created by TBL. So, although it could be possible to convert transformation rules automatically to LT notation, human intervention is still advisable to make them more general.

In all cases we observed the same phenomenon as with the test results above in section 4: we were able to enhance the recall of many existing confused words rules in LT by using TBL, but the precision of some of the automatically-acquired rules was too low. Here are some examples of rules acquired using *µ*-TBL toolkit (Lager 1999) in its Prolog-style notation:

```
wd:they>the   <- tag:'DT'@[5] o
wd:they>the   <- tag:'NN'@[-2] o
wd:they>the   <- tag:'JJ,NN,RB'@[1] o
wd:they>the   <- tag:'NNS,VBZ'@[5] o
wd:there>their <- wd:of@[-1] o
wd:they>the   <- tag:'JJ,NN'@[1] o
wd:there>their <- wd:'SENT_START'@[3] o
wd:they>the   <- tag:'NN'@[1] o
wd:they>the   <- tag:'SENT_START'@[-3] o
```

---

6 We accepted all possible candidates for rules from the ranked list; we could apply a threshold to filter out the rules that did not have sufficient examples to learn their features from. This would make recall probably lower but could enhance precision.

As can be seen in the example, we experimented with rule templates quite freely, and tried to see what would be the result of using the feature which is the 5[th] one after the current one. Contrary to our expectations, the results in most cases are still somewhat useful, although it is of course safer not to use such controversial templates.

The input text was tagged with the internal LanguageTool POS tagger (which disambiguates only to a very limited degree due to specific requirements of grammar checking) to create rules that may be almost directly translated into LT notation with limited manual changes. Polish and English automatically-acquired rules both seem to be useful and are gradually implemented, while human intervention is limited to enhancing rule precision.

In the future, we want to port the current prototype framework (consisting mostly of AWK scripts and Unix tools) to Java, including bootstraping the error corpus from Wikipedia history dump (see Miłkowski 2008). This would offer more platform-independence. This way, the process of adding a new language to LanguageTool would be greatly facilitated, which is one of the practical goals of the project. We consider also using one of the Java implementations of TBL learning directly in our learning tool, and using automatic translation from TBL rules to LT formalism.

We think that even though automatically-acquired rules not always have a satisfactory level of precision, TBL may actually help with development of rules. If minimizing manual intervention were our goal, we could simply use only the best ranked rules with the best precision and test it on additional corpora to make sure they do not raise an unusually high number of alarms, at the price of lower recall. The best automatically-acquired rules are sometimes simpler than the ones proposed by human linguists, and may become a source of inspiration. Therefore, though we cannot completely rely on the automatically generated results, they seem to alleviate creation of rules in a significant way.